\pdfoutput=1
\documentclass[letterpaper, 10 pt, conference]{ieeeconf}

\IEEEoverridecommandlockouts
\usepackage{graphicx}
\usepackage{amsmath}
\usepackage{amssymb}
\usepackage{bm}
\usepackage{booktabs}
\usepackage{multirow}
\usepackage{cite}
\usepackage{url}
\usepackage{hyperref} 
\usepackage[caption=false,font=footnotesize]{subfig}

\title{\LARGE \bf
Dynamic-LIVO: A Dynamic-Aware LiDAR-Inertial-Visual Odometry System Using Spatio-Temporal Normals
}

\author{Zhixin Zhang$^{1}$, Samuel Ahiwe$^{2}$, Matthew Hale$^{1}$, Liang Zhao$^{3}$, Pawel Ladosz$^{2}$
\thanks{This work was partially funded by the Robotics and AI Collaboration (RAICo).}
\thanks{$^{1}$Zhixin Zhang, Matthew Hale are with the Department of Electrical and Electronic Engineering, University of Manchester, Manchester M13 9PL, U.K.
Email: {\tt\small zhixin.zhang@manchester.ac.uk}}
\thanks{$^{3}$Samuel Ahiwe, Pawel Ladosz are with the Department of Mechanical, Aerospace and Civil Engineering, University of Manchester, Manchester M13 9PL, U.K.
Email: {\tt\small pawel.ladosz@manchester.ac.uk}}
\thanks{$^{2}$Liang Zhao is with the School of Informatics, University of Edinburgh, Edinburgh EH8 9AB, U.K.
Email: {\tt\small liang.zhao@ed.ac.uk}}
}

\begin{document}

\maketitle
\thispagestyle{empty}
\pagestyle{empty}

\begin{abstract}
This paper proposes Dynamic-LIVO, a dynamic-aware LiDAR-Inertial-Visual Odometry (LIVO) system for robust state estimation and static colored mapping in dynamic environments. Dynamic-LIVO employs Spatio-Temporal (S-T) normal analysis to identify dynamic LiDAR points and propagates the resulting classification to both LiDAR-inertial and visual-inertial updates, preventing dynamic LiDAR measurements and their associated visual observations from affecting state estimation and mapping. However, S-T normal estimation can be unreliable in newly observed and spatially sparse regions due to insufficient spatio-temporal observations. To address this issue, we introduce a time-delayed S-T normal estimation strategy that defers the classification of insufficiently constrained points and re-evaluates them as additional observations become available. This strategy improves dynamic classification reliability while preserving valid static points for map construction. Extensive experiments on public and self-collected datasets with diverse sensor configurations demonstrate that Dynamic-LIVO improves localization accuracy and produces cleaner static colored maps in challenging dynamic environments. The source code and self-collected dataset will be publicly released upon acceptance.
\end{abstract}

\section{INTRODUCTION}
Simultaneous localization and mapping (SLAM) is a fundamental technology for autonomous robots, enabling accurate state estimation and 3D map reconstruction in unknown environments. Based on the primary sensing modality, existing SLAM approaches can be broadly categorized into vision-based~\cite{mur2017orb,qin2018vins} and LiDAR-based~\cite{shan2020lio,xu2022fastlio2} methods, each with its own complementary strengths and limitations. Vision-based methods exploit rich texture information from camera measurements, but their performance can degrade in low-texture or poor-illumination environments. In contrast, LiDAR-based methods directly provide accurate depth measurements and are insensitive to illumination, but may suffer from degeneracy in environments with insufficient geometric constraints. These limitations can degrade system performance when relying on sensing modality alone.

To leverage the complementary strengths of vision and LiDAR, recent studies have increasingly explored the fusion of visual and LiDAR measurements, leading to the development of LiDAR-inertial-visual odometry (LIVO) systems with significant performance improvements. LVI-SAM~\cite{shan2021lvi} proposed a tightly coupled LiDAR-visual-inertial framework that integrates LiDAR-inertial and visual-inertial odometry within a factor graph. R3LIVE~\cite{lin2022r} further integrated LiDAR, inertial, and visual measurements within an Iterated Error-State Kalman Filter (IESKF)~\cite{bell1993iterated} framework to enable photorealistic colored mapping. More recently, FAST-LIVO2~\cite{zheng2024fast} introduced an efficient direct LiDAR-inertial-visual fusion framework that jointly exploits geometric and photometric information, significantly improving both estimation accuracy and computational efficiency.

\begin{figure}
    \centering
    \includegraphics[width=0.95\linewidth]{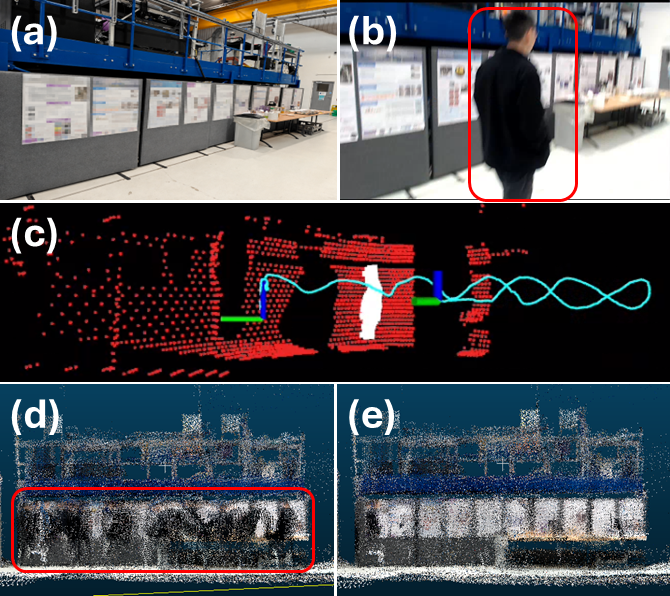}
    \caption{Dynamic-LIVO results on the self-collected PosterWall sequence. (a) Test environment. 
    (b) Example image showing a moving person during data collection. 
    (c) Dynamic filtering result for a representative LiDAR scan, where white and red points denote filtered dynamic points and retained static points, respectively. 
    (d)(e) Colored mapping results of FAST-LIVO2 and Dynamic-LIVO, respectively. The red rectangle in (d) highlight noticeable traces of moving objects retained in the FAST-LIVO2 map, which are effectively removed by Dynamic-LIVO in (e).}
    \label{fig:firstfigure}
\end{figure}

However, existing LIVO methods generally rely on the static-world assumption, without explicitly accounting for moving objects. Dynamic objects can introduce erroneous geometric and visual measurements into state estimation and mapping, thereby degrading localization accuracy and mapping quality. Since many LIVO systems rely on LiDAR measurements to provide depth information for visual observations, dynamic LiDAR measurements can affect not only LiDAR-inertial estimation but also subsequent visual updates. Therefore, effectively identifying and removing dynamic LiDAR points is important for robust LIVO in dynamic environments. In the field of LiDAR-inertial odometry (LIO), various dynamic-aware methods~\cite{yuan2025lidar,lichtenfeld2024efficient,jia2025trlo} have been proposed to detect and remove dynamic points from LiDAR measurements. Among them, Spatio-Temporal (S-T) normal analysis~\cite{falque2023dynamic,le2024real,chen2025breaking} has emerged as an effective approach for identifying dynamic points by jointly exploiting spatial geometry and temporal variation. In this work, we adapt S-T normal-based dynamic detection to the LIVO framework, enabling dynamic measurements to be excluded from both state estimation and colored map construction for cleaner static environment reconstruction.

Despite its effectiveness, S-T normal analysis can become unreliable in newly observed and spatially sparse regions, where insufficient spatio-temporal support may lead to inaccurate S-T normal estimation and erroneous dynamic classification. Such misclassification may not only introduce dynamic measurements into state estimation but also incorrectly discard valid static points, degrading the completeness and quality of the constructed static map. To address this limitation, we propose a time-delayed S-T normal estimation strategy that defers the classification of insufficiently constrained points until additional observations become available. By exploiting subsequently accumulated spatio-temporal information, these points are re-evaluated with more reliable S-T normals, reducing classification errors and preserving valid static observations for static map construction. The main contributions of this work are summarized as follows:
\begin{itemize}

\item We propose Dynamic-LIVO, a learning-free dynamic-aware LIVO system for robust state estimation and clean static colored mapping in dynamic environments.

\item We employ S-T normal analysis to detect dynamic LiDAR points from raw scans and propagate the resulting dynamic classification to both LIO and VIO subsystems, preventing dynamic LiDAR measurements and their associated visual observations from contributing to state estimation and mapping.

\item We propose a time-delayed S-T normal estimation strategy that defers the classification of points in degenerate cases until sufficient observations are accumulated, improving dynamic classification reliability and recovering valid static points for map construction.

\item Extensive experiments with diverse sensor configurations demonstrate that Dynamic-LIVO improves localization accuracy and produces cleaner static colored maps in challenging dynamic environments.

\end{itemize}
\section{RELATED WORK}

\subsection{LiDAR-Inertial-Visual SLAM}
Recent years have witnessed rapid development in LiDAR-inertial-visual odometry (LIVO), driven by the complementary characteristics of LiDAR, visual sensors and improved computation available. LIC-Fusion~\cite{zuo2019lic,zuo2020lic} proposed a tightly coupled framework that integrates IMU measurements, visual features, and two types of LiDAR features within an MSCKF framework. LVI-SAM~\cite{shan2021lvi} further combined LiDAR-inertial and visual-inertial odometry in a factor graph framework, where visual features were associated with LiDAR-derived depth measurements. R3LIVE~\cite{lin2022r,lin2024r} further integrated LiDAR and visual measurements within an IESKF framework by jointly minimizing LiDAR and visual reprojection errors, enabling robust state estimation and real-time colored mapping.

Despite their effectiveness, these methods rely heavily on feature extraction and feature association from both LiDAR and visual measurements, which inevitably increase computational complexity and limit real-time performance. To address this limitation, FAST-LIVO2~\cite{zheng2022fast,zheng2024fast} introduced a direct LiDAR-inertial-visual fusion framework that eliminates the need for explicit feature extraction and association, thereby significantly improving computational efficiency. Building upon the efficient direct fusion framework of FAST-LIVO2, we extend the system to dynamic environments by incorporating S-T normal analysis to identify and remove dynamic LiDAR points. During depth association, only static LiDAR points are used to provide depth information for visual features, while image points corresponding to dynamic regions are excluded from subsequent visual updates.

\subsection{Dynamic-Aware LIO and LIVO}
LIO and LIVO in dynamic environments have attracted increasing attention, as moving objects violate the static-world assumption and introduce erroneous measurements into state estimation and mapping. Learning-based methods~\cite{chen2019suma++,henein2020dynamic,shi2024dynam,Schmid2024Khronos} identify dynamic objects using semantic segmentation or object detection and exclude their associated measurements from state estimation. However, their performance depends on predefined semantic categories and detection accuracy, which may limit generalization to unseen objects while introducing additional computational overhead.

In contrast, learning-free methods detect dynamic measurements based on geometric or temporal information without relying on predefined semantic categories~\cite{postica2016robust,lichtenfeld2024efficient,yuan2025lidar}. Among these approaches, S-T normal-based methods~\cite{falque2023dynamic,le2024real,chen2025breaking} have attracted increasing attention due to their low computational cost and strong generalizability. By jointly analyzing the spatial and temporal distribution of 3D points, these methods identify measurements exhibiting motion-induced spatio-temporal inconsistencies. However, existing S-T normal-based approaches have primarily been developed within LIO frameworks, where dynamic measurements are handled only in the LiDAR--inertial estimation pipeline. Moreover, reliable S-T normal estimation requires sufficient observations across both spatial and temporal domains, which may not be available in newly observed or spatially sparse regions. To address these limitations, Dynamic-LIVO extends S-T normal-based dynamic point detection to the LIVO framework and introduces a time-delayed estimation strategy that postpones the classification of such points until sufficient spatio-temporal observations become available.

\begin{figure}[t]
\vspace{2mm}
    \centering
    \includegraphics[width= 0.45\textwidth]{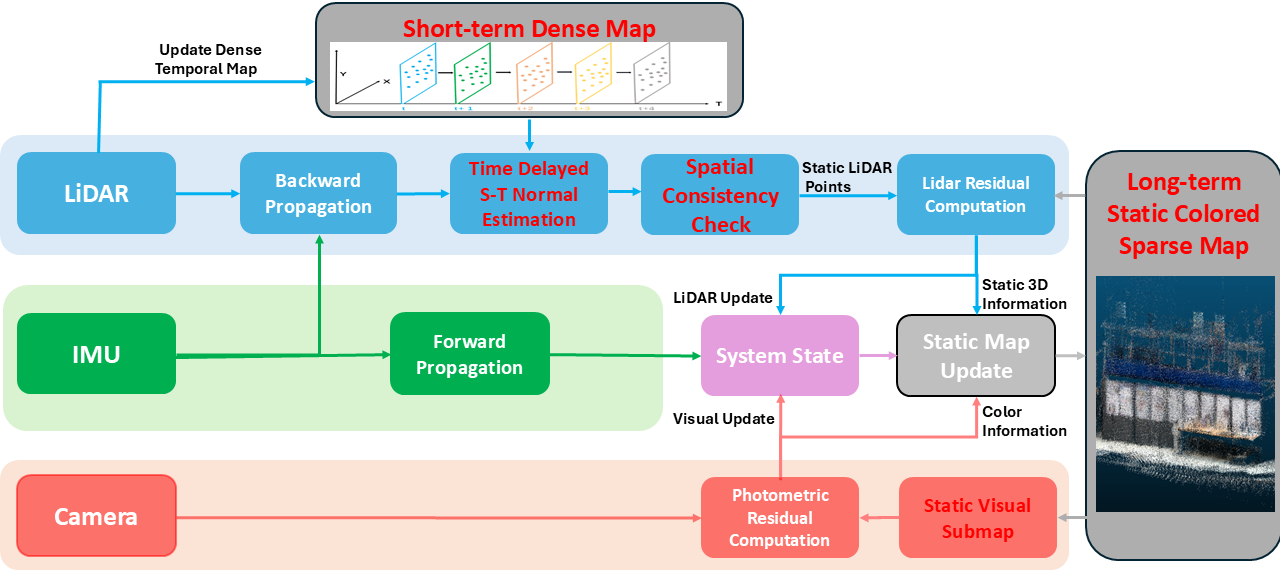}
    \caption{Overview of the Dynamic-LIVO framework. The system consists of four main components, represented by different color-coded boxes. Modules highlighted with red text denote the components newly introduced in Dynamic-LIVO.} 
    \label{fig:framework}
\end{figure}

\section{METHODOLOGY}
\subsection{System Overview}
Built upon FAST-LIVO2~\cite{zheng2024fast}, Dynamic-LIVO extends the original framework with dynamic-aware processing for robust state estimation and static colored mapping. The system consists of four main components: LiDAR processing, IMU propagation, visual processing, and static colored mapping. The first three components perform state estimation within an IESKF framework, while the estimated odometry is used for static colored map construction. In addition, a short-term dense map is maintained for S-T normal-based dynamic point detection. For each LiDAR scan, dynamic points are first identified and removed, and the remaining static points are used for LIO state estimation and long-term map update. The resulting static map is further used to provide depth information for visual observations in the VIO update.

\begin{figure*}[t]
    \centering
    \subfloat[1D Example of S-T Normal]{
        \includegraphics[width=0.3\textwidth]{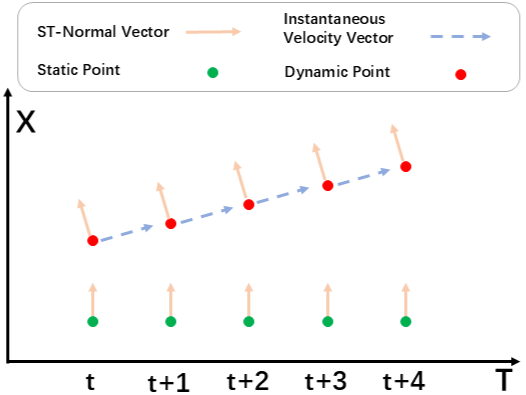}}
        \hspace{0.02\textwidth}
    \subfloat[2D Example of Static S-T Normal]{
        \includegraphics[width=0.3\textwidth]{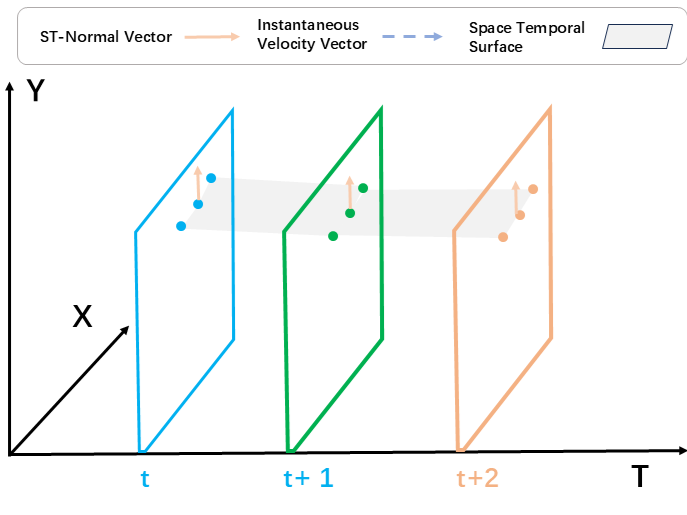}}
        \hspace{0.02\textwidth}
    \subfloat[2D Example of Dynamic S-T Normal]{
        \includegraphics[width=0.3\textwidth]{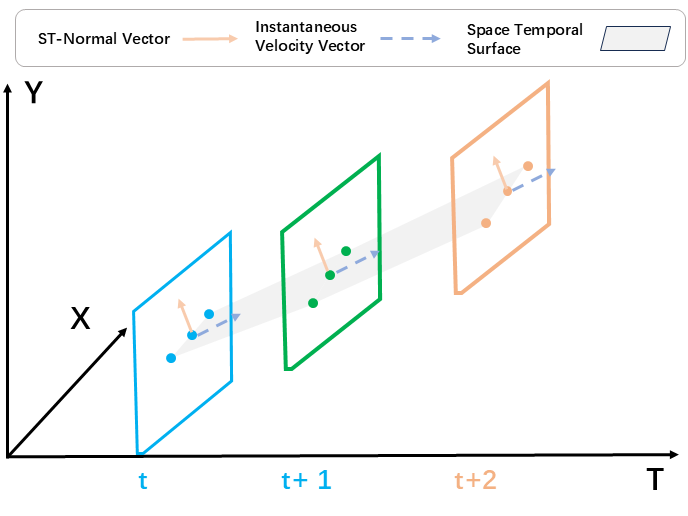}}
    \caption{Illustration of S-T normal analysis in 1D and 2D spatial domains.
    (a) Green and red points represent static and dynamic observations, respectively, from timestamps \(t\) to \(t+4\). The S-T normal of the static observations is orthogonal to the temporal axis, resulting in a zero temporal component, whereas that of the dynamic observations exhibits a non-zero temporal component.
    (b) Three static points observed from \(t\) to \(t+2\) form the gray rectangular S-T surface. Its S-T normal is orthogonal to the temporal axis, resulting in a zero temporal component.
    (c) Three dynamic points observed from \(t\) to \(t+2\) form the gray inclined S-T surface. Its S-T normal is no longer orthogonal to the temporal axis, resulting in a non-zero temporal component.}
    \label{fig:S-Tnormal}
\end{figure*}

\subsection{Dynamic Point Detection}
In this subsection, we first introduce the principle of S-T normal-based dynamic point detection and then present the improvements incorporated into Dynamic-LIVO over existing S-T normal-based methods~\cite{falque2023dynamic,le2024real,chen2025breaking}.

\subsubsection{S-T Normal Analysis}
For a point cloud $\mathbf{P}^{j}$ captured at timestamp $t_j$, each point can be represented in the 4D spatio-temporal (S-T) domain as
\begin{equation}
\tilde{\mathbf{p}}_{i}^{j}
=
\left\{
\mathbf{p}_{i}^{j}, t_j
\mid
\mathbf{p}_{i}^{j} \in \mathbb{R}^{3},
t_j \in \mathbb{R}
\right\}.
\end{equation}
where $\mathbf{p}_{i}^{j}=[x_i^j,y_i^j,z_i^j]^{\mathrm{T}}$ denotes the $i$-th 3D point in the $j$-th LiDAR frame. 

As a local 3D spatial surface evolves over time, its observations across consecutive timestamps form an S-T surface $S$ in the 4D spatio-temporal domain. This S-T surface can be represented by an implicit function as
\begin{equation}
S
=
\left\{
\tilde{\mathbf{p}}
=
[\mathbf{p}^{T}, t]^{T}
\in \mathbb{R}^{4}
\mid
F(\mathbf{p},t)=0
\right\}.
\end{equation}

By differentiating both side of the implicit function $F(\mathbf{p}(t),t)=0$ with respect to time, we obtain
\begin{equation}
\frac{dF}{dt}
=
\frac{\partial F}{\partial x}\frac{dx}{dt}
+
\frac{\partial F}{\partial y}\frac{dy}{dt}
+
\frac{\partial F}{\partial z}\frac{dz}{dt}
+
\frac{\partial F}{\partial t}
= 0.
\end{equation}
The first three derivatives form the spatial component of the S-T normal,
$\mathbf{n}_{s} =
[\frac{\partial F}{\partial x},
\frac{\partial F}{\partial y},
\frac{\partial F}{\partial z}]^{T}$,
while $n_t=\frac{\partial F}{\partial t}$ denotes its temporal component. Since
$\mathbf{v}=[\frac{dx}{dt},\frac{dy}{dt},\frac{dz}{dt}]^{T}
=[v_x,v_y,v_z]^{T}$
represents the velocity of the surface in the spatial domain, the above equation can be rewritten as
\begin{equation}
\mathbf{n}_{s}^{T}\mathbf{v} + n_t = 0.
\end{equation}
This relationship indicates that the temporal component of the S-T normal contains information related to the motion of the observed surface. For a static point, where $\mathbf{v}=\mathbf{0}$, the temporal component satisfies $n_t=0$. In contrast, for a moving point with a non-zero velocity component along the spatial normal direction, $n_t$ becomes non-zero. Fig.~\ref{fig:S-Tnormal} illustrates this relationship through 1D and 2D S-T normal examples. Therefore, the S-T normal can be used to distinguish between dynamic and static points.

\subsubsection{S-T Normal Estimation}
To estimate the S-T normal of a 4D point $\tilde{\mathbf{p}}_{i}^{j}$, a set of neighboring points $\mathcal{N}_{i}^{j}$ is searched in the 4D S-T domain to fit a local tangent hyperplane. The eigenvector corresponding to the smallest eigenvalue represents the normal direction of the fitted hyperplane and is therefore taken as the S-T normal vector of $\tilde{\mathbf{p}}_{i}^{j}$.

\begin{equation}
\operatorname{cov}_{i}^{j}
=
\frac{1}{|\mathcal{N}_{i}^{j}|}
\sum_{\mathbf{p}_{u}^{v} \in \mathcal{N}_{i}^{j}}
\left(
\begin{bmatrix}
\mathbf{p}_{u}^{v} \\
t_{u}^{v}
\end{bmatrix}
-
\mathbf{m}_{i}^{j}
\right)
\left(
\begin{bmatrix}
\mathbf{p}_{u}^{v} \\
t_{u}^{v}
\end{bmatrix}
-
\mathbf{m}_{i}^{j}
\right)^{\mathrm{T}}.
\end{equation}
where $\mathbf{m}_{i}^{j}$ is the mean vector:
\begin{equation}
\mathbf{m}_{i}^{j}
=
\frac{1}{|\mathcal{N}_{i}^{j}|}
\sum_{\mathbf{p}_{u}^{v} \in \mathcal{N}_{i}^{j}}
\begin{bmatrix}
\mathbf{p}_{u}^{v} \\
t_{u}^{v}
\end{bmatrix}.
\end{equation}

Accurate S-T normal estimation requires a sufficiently dense local neighborhood in the 4D S-T domain. However, the long-term map is typically downsampled to reduce memory consumption and computational cost, resulting in insufficient local point density for reliable S-T normal estimation. To address this issue, Dynamic-LIVO maintains an additional short-term dense map dedicated to S-T normal estimation. This map preserves recent LiDAR measurements without the downsampling and dynamic point filtering applied to the long-term map. Specifically, a sliding temporal window is employed to retain the recent 1s of LiDAR measurements, providing sufficiently dense neighborhoods for reliable S-T normal computation.

\subsubsection{Time-Delayed S-T Normal Estimation}
\begin{figure}
    \centering
    \subfloat[Newly Observed Region]{
        \includegraphics[width=0.23\textwidth]{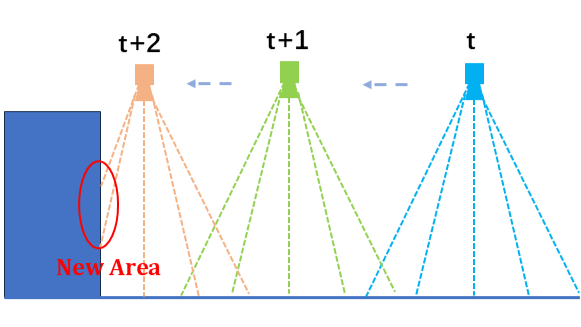}}
    \subfloat[Spatially Sparse Region]{
        \includegraphics[width=0.23\textwidth]{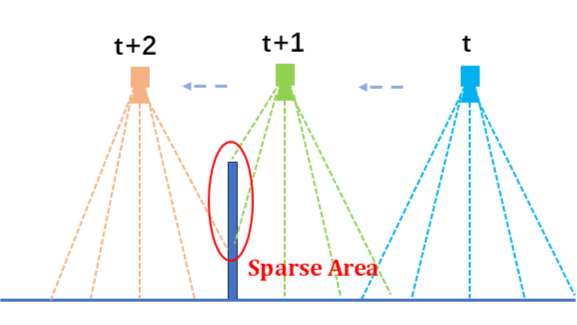}}
    \caption{Illustration of two challenging cases for S-T normal estimation. 
    (a) A newly observed region lacks sufficient temporal observations. 
    (b) A spatially sparse region contains insufficient neighboring points for reliable S-T surface fitting.}
    \label{fig:degenerate}
\end{figure}

Despite the timestamp-aware neighbor selection strategy, reliable S-T normal estimation remains challenging in newly observed or spatially sparse regions, as shown in Fig.~\ref{fig:degenerate}. When a point is first observed, its local neighborhood may lack sufficient spatial and temporal observations to reliably fit the local S-T surface. Consequently, the estimated S-T normal may contain an erroneous temporal component, leading to incorrect dynamic classification. A similar issue arises in spatially sparse regions, where the neighboring observations are distributed over a large spatial extent and therefore cannot reliably characterize the local S-T surface.

To address these limitations, we introduce a time-delayed S-T normal estimation strategy, which postpones the final S-T classification when sufficient spatio-temporal support is unavailable. For each query point, the $K=20$ nearest neighbors are first retrieved according to their spatial distances. The neighborhood is considered invalid if fewer than $K$ neighbors are found, if the distance between the query point and its farthest selected neighbor exceeds a predefined spatial threshold, or if the selected neighbors do not provide sufficient temporal diversity. Specifically, the neighboring points are grouped into temporal bins, and a valid neighborhood is required to span at least three temporal bins while preventing a single bin from dominating the neighborhood. Points that fail any of these conditions are labeled as \textit{unknown}, since their local observations are insufficient for reliable S-T normal estimation.

Instead of immediately assigning these unknown points as static or dynamic, Dynamic-LIVO temporarily retains them for delayed re-evaluation. As subsequent LiDAR frames are incorporated into the short-term dense map, the retained points are re-evaluated using the updated local neighborhood. Once sufficient spatial and temporal support becomes available, the S-T normal is estimated and the point is classified as stable or unstable according to its temporal component. Otherwise, the point remains \textit{unknown} and is retained for further evaluation in subsequent frames. If it still cannot be reliably classified after six retry attempts, this unknown point is discarded and is not inserted into the static map.

\subsubsection{Spatial Consistency Check}
Although S-T normal analysis provides an effective temporal cue for distinguishing dynamic and static observations, the initial results may still contain errors, particularly false positives. To further refine these results,inspired by the spatial consistency strategy in BTSA~\cite{chen2025breaking}, Dynamic-LIVO performs a Spatial Consistency Check (SCC) based on the spatial distribution of the initial S-T normal-based classification. Specifically, the detected unstable points are first upsampled by retrieving their neighboring points from the original dense LiDAR scan. DBSCAN clustering~\cite{ester1996density} is then applied to remove isolated false positives and group spatially consistent points into candidate dynamic clusters. Finally, these candidate clusters are compared with the short-term static voxel map through a spatial overlap check. Clusters with sufficient overlap with previously observed static regions are rejected as false positives, while the remaining clusters are retained as the final dynamic observations.

\subsection{Dynamic-Aware State Estimation}
Based on the dynamic classification results obtained from the S-T normal analysis and subsequent spatial consistency check, Dynamic-LIVO further incorporates dynamic awareness into both LIO and VIO subsystem state estimation. The key principle is to prevent measurements associated with dynamic objects from contributing to the state update, thereby reducing the adverse influence of independently moving objects on pose estimation.

\subsubsection{Dynamic-Aware LIO Subsystem}
For each LiDAR scan, IMU measurements are first used for state propagation and motion compensation, transforming the LiDAR points acquired at different timestamps into a common reference frame to obtain a motion-compensated scan. The reconstructed scan is subsequently processed by the proposed dynamic filtering module, where points associated with dynamic objects are identified and excluded from the LiDAR state update. The remaining static points are then associated with the long-term static map to construct geometric constraints for state estimation. Specifically, for each static LiDAR point, neighboring points are retrieved from the long-term static map and used to fit a local planar surface. The corresponding point-to-plane residual is then constructed as the geometric constraint for the LIO state update, and the state is iteratively refined through the IESKF by minimizing these residuals. By constructing the LiDAR measurement update exclusively from static observations, the influence of independently moving objects on state estimation is effectively suppressed.

After the LIO state update, the filtered static scan is transformed into the world coordinate frame using the updated state for long-term static map construction. To generate a colored static map, the most recent image observation is further used to assign color information to the LiDAR points through LiDAR--camera projection. The resulting colored static points are then integrated into the long-term map, yielding a continuously updated static representation of the environment.

\subsubsection{Dynamic-Aware VIO Subsystem}
The VIO subsystem requires LiDAR measurements to provide depth information for image patches used in the visual state update. In FAST-LIVO2, the most recent LiDAR scan is directly used as a submap for depth association without explicitly considering dynamic objects. Consequently, LiDAR points belonging to moving objects may provide depth information to the corresponding image regions, allowing dynamic visual observations to participate in the subsequent state update.

To address this issue, Dynamic-LIVO propagates the LiDAR-based dynamic classification to the visual subsystem through LiDAR--image projection. The dynamically filtered LiDAR scan is used to construct the visual submap and provide depth information for image patches. If a LiDAR point projected onto an image patch is identified as dynamic, the corresponding patch is regarded as a dynamic-associated visual observation and is excluded from both the visual map construction and the subsequent state update. Consequently, only image patches associated with static LiDAR measurements are retained for visual state estimation, preventing observations of moving objects from introducing erroneous photometric constraints.

The retained image patches with valid 3D correspondences are projected onto the current image according to the estimated state, and photometric residuals are computed from the intensity differences between the reference and current patches. These residuals are subsequently used to iteratively update the system state. Further details of the visual state estimation are provided in the original FAST-LIVO2 work~\cite{zheng2024fast}.


\section{EXPERIMENTS}
\begin{figure}[t]
    \centering
    \includegraphics[width= 1.0\linewidth]{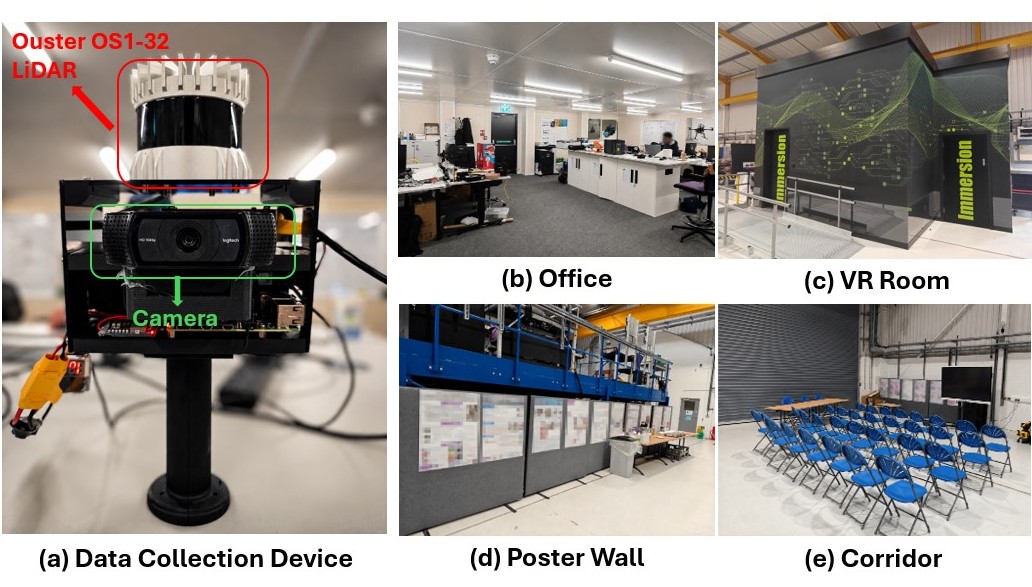}
    \caption{Self-collected dataset. (a) Data collection platform. (b)--(e) Test scenarios.}
    \label{fig:SelfCollect}
\end{figure}

We evaluated Dynamic-LIVO on several challenging dynamic sequences from both the publicly available M3DGR~\cite{zhang2025towards} dataset and our self-collected dataset. These sequences cover diverse scenarios and different LiDAR sensing configurations. The M3DGR sequences were recorded using two different solid-state LiDAR sensors, namely the Livox Mid-360 and Livox Avia, which exhibit substantially different sensing fields of view (FoVs). The Mid-360 provides a $360^\circ$ horizontal FoV with a vertical FoV ranging from $-7^\circ$ to $52^\circ$, whereas the Avia has a narrower $70.4^\circ$ horizontal FoV and a $77.2^\circ$ vertical FoV. Both LiDAR sensors operate at 10~Hz and are equipped with built-in 6-axis IMUs operating at 200~Hz. An Intel RealSense D435i camera was used to provide visual measurements.
Our self-collected sequences were recorded using an Ouster OS1-32 spinning LiDAR operating at 10~Hz, with its built-in IMU operating at 100~Hz, together with a Logitech C920 camera operating at 30~Hz, as shown in Fig.~\ref{fig:SelfCollect}(a). The self-collected dataset covers four different test scenarios, as shown in Fig.~\ref{fig:SelfCollect}(b)--(e), with people continuously moving through the environments to create challenging dynamic conditions during data collection. Therefore, the evaluation covers both solid-state and spinning LiDAR configurations. 

We evaluated the proposed Dynamic-LIVO in terms of localization accuracy, static colored mapping quality, and computational efficiency. All experiments were conducted on a computer equipped with an Intel Core Ultra 9 CPU, 32 GB of RAM, and an NVIDIA RTX PRO 3000 GPU with 12 GB of VRAM, running Ubuntu 20.04.

\subsection{Localization Accuracy}

\begin{table*}[t]
\vspace{2mm}
    \centering
    \caption{Comparison of absolute trajectory error (ATE) RMSE (m) on the M3DGR dataset.}
    \small
    \setlength{\tabcolsep}{5pt}
    \label{tab::m3dgr}
    \resizebox{0.85 \textwidth}{!}{
    \begin{tabular}{lcccccc}
        \toprule
        \multirow{2}{*}{\textbf{Method}} 
        & \multicolumn{4}{c}{\textbf{Livox Mid-360}}
        & \multicolumn{2}{c}{\textbf{Livox Avia}} \\
        
        \cmidrule(lr){2-5}
        \cmidrule(lr){6-7}
        
        & \textbf{Dynamic01}
        & \textbf{Dynamic02}
        & \textbf{Dynamic03}
        & \textbf{Dynamic04}
        & \textbf{Dynamic03}
        & \textbf{Dynamic04} \\
        
        \midrule
        
        R3LIVE~\cite{lin2022r}
        & 1.102 & 1.067 & 1.196 & 1.502
        & 0.900 & 0.510 \\
        
        LVI-SAM~\cite{shan2021lvi}
        & 1.298 & 1.193 & 12.544 & 24.962
        & 1.281 & 5.341 \\
        
        FAST-LIVO~\cite{zheng2022fast}
        & FAIL & FAIL & FAIL & FAIL 
        & 1.249 & 0.949 \\
        
        FAST-LIVO2~\cite{zheng2024fast}
        & \underline{0.149} & \underline{0.145} & \underline{0.158} & 0.204 
        & \textbf{0.117} & \underline{0.100} \\
        
        SR-LIVO~\cite{yuan2024sr}
        & FAIL & FAIL & FAIL & FAIL
        & 0.124 & FAIL \\
        
        \midrule
        
        BTSA~\cite{chen2025breaking}
        & \underline{0.149} & 0.149 & 0.186 & \textbf{0.202}
        & 0.147 & 0.128 \\
        
        \midrule
        
        \textbf{Dynamic-LIVO} 
        & \textbf{0.142} & \textbf{0.132} & \textbf{0.145} & \textbf{0.202}
        & \textbf{0.117} & \textbf{0.094} \\
        
        \bottomrule
    \end{tabular}
    }

    \vspace{1mm}
    \parbox{\linewidth}{\footnotesize
    \raggedright
    * The best results are shown in \textbf{bold}, and the second-best results are \underline{underlined}, respectively.}
\end{table*}

To evaluate localization accuracy, we compared the proposed Dynamic-LIVO with several state-of-the-art (SOTA) LIVO methods primarily designed for static environments, including R3LIVE~\cite{lin2022r}, LVI-SAM~\cite{shan2021lvi}, FAST-LIVO~\cite{zheng2022fast}, FAST-LIVO2~\cite{zheng2024fast}, and SR-LIVO~\cite{yuan2024sr}. In addition, BTSA~\cite{chen2025breaking}, an LIO method specifically designed for dynamic environments, was included for comparison.

\subsubsection{M3DGR} For localization evaluation, we first selected four M3DGR sequences containing dynamic objects, including two indoor and two outdoor sequences. These sequences involve representative dynamic objects, such as moving pedestrians and vehicles. Ground-truth trajectories are available, and the root mean square error (RMSE) of the absolute trajectory error (ATE) was computed using the evo~\cite{grupp2017evo} evaluation tool. To ensure a fair comparison, all evaluated methods were configured using the sensor calibration and configuration parameters provided by the M3DGR dataset.

Table~\ref{tab::m3dgr} presents the localization results of all baseline methods and Dynamic-LIVO on the M3DGR dataset across different dynamic sequences and LiDAR configurations. Dynamic-LIVO achieves the best localization accuracy under the Livox Mid-360 configuration, ranking first on all four dynamic sequences. In particular, Dynamic-LIVO consistently outperforms FAST-LIVO2, demonstrating that the proposed dynamic point filtering strategy can effectively reduce the adverse influence of moving objects on state estimation while preserving sufficient static geometric information for accurate localization.

Under the Livox Avia configuration, we report results only on Dynamic03 and Dynamic04. Dynamic01 and Dynamic02 are excluded from the quantitative comparison because the LiDAR scans become temporarily unavailable during these sequences, resulting in a prolonged loss of effective LiDAR measurements and causing the evaluated LIO/LIVO methods to fail. On the remaining two sequences, Dynamic-LIVO achieves the best localization accuracy, demonstrating robust performance despite the substantially smaller FoV.

\begin{table}[t]
    \centering
    \caption{Comparison of End-to-start position error (m) on the self-collected dataset.}
    \label{tab:self-collect}
    \begin{tabular}{c|cccc}
        \toprule
        \textbf{Method}
        & \textbf{Office}
        & \textbf{VR-Room}
        & \textbf{Poster Wall}
        & \textbf{Corridor} \\
        \midrule
        
        R3LIVE~\cite{lin2022r}
        & 0.257 & 0.107 & 0.029 & 0.088 \\
        
        LVI-SAM~\cite{shan2021lvi}
        & 1.977 & 0.078 & FAIL & FAIL \\
        
        FAST-LIVO~\cite{zheng2022fast}
        & 1.025 & 0.731 & 4.408 & 1.049 \\
        
        FAST-LIVO2~\cite{zheng2024fast}
        & 0.174 &  \underline{0.023} & \underline{0.020} & 0.061 \\
        
        SR-LIVO~\cite{yuan2024sr}
        & \textbf{0.132} & 0.696 & 0.025 & 0.092 \\

        \midrule
        
        BTSA~\cite{chen2025breaking}
        & 0.192 & 0.580 & 0.022 & \underline{0.038} \\
        
        \midrule
        
        \textbf{Dynamic-LIVO}
        & \underline{0.171} & \textbf{0.020} & \textbf{0.016} &  \textbf{0.032}\\
        
        \bottomrule
    \end{tabular}

    \vspace{1mm}
    \parbox{\linewidth}{\footnotesize
    \raggedright
    \raggedright
    * The best results are shown in \textbf{bold}, and the second-best results are \underline{underlined}, respectively.}
\end{table}

\subsubsection{Self-Collected Datasets}

Since ground-truth trajectories are unavailable for our self-collected sequences, each sequence was recorded such that the sensor platform started and ended at the same location. In addition, the sensor platform was kept stationary for the first 2s of each sequence to ensure proper initialization of all evaluated methods. Localization accuracy was therefore evaluated using the end-to-start position error.

Table~\ref{tab:self-collect} presents the end-to-start position errors of all baseline methods and Dynamic-LIVO on our self-collected dataset. Dynamic-LIVO achieves the best overall localization performance on the self-collected dataset, ranking first in three out of the four sequences and second in the remaining sequence. Moreover, Dynamic-LIVO consistently outperforms its baseline, FAST-LIVO2, across all four sequences. This improvement demonstrates the effectiveness of the proposed spatio-temporal normal-based dynamic point filtering strategy in mitigating the adverse effects of dynamic objects on state estimation, thereby improving localization accuracy in dynamic environments. Compared with BTSA, which also employs spatio-temporal normal analysis for dynamic point detection but relies on LiDAR-inertial measurements, Dynamic-LIVO achieves better overall localization performance by further incorporating visual information, providing complementary constraints in dynamic environments.

\subsection{Static Colored Mapping}
We further evaluate the static colored mapping performance of Dynamic-LIVO to assess the effectiveness of dynamic point filtering and its dense colored mapping capability. Specifically, we compare the colored maps generated by FAST-LIVO2 and Dynamic-LIVO across different dynamic environments.

\subsubsection{M3DGR}
\begin{figure}
    \centering
    \subfloat[Real-time colored mapping results  of FAST-LIVO2 (left) and Dynamic-LIVO (right) on Dynamic01 using Livox Mid-360. The red circle highlights several pedestrians walking inside the room.]{
    \includegraphics[width=1.0\linewidth]{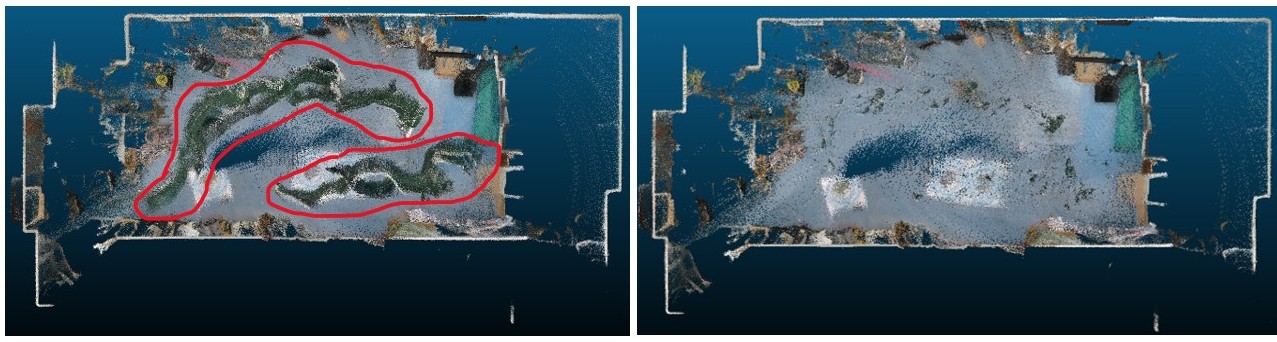}}
    \\ 
    \subfloat[Real-time colored mapping results of FAST-LIVO2 (left) and Dynamic-LIVO (right) on Dynamic03 using Livox Mid-360. Two representative regions are enlarged to highlight the differences in dynamic point removal.]{\includegraphics[width= 1.0\linewidth]{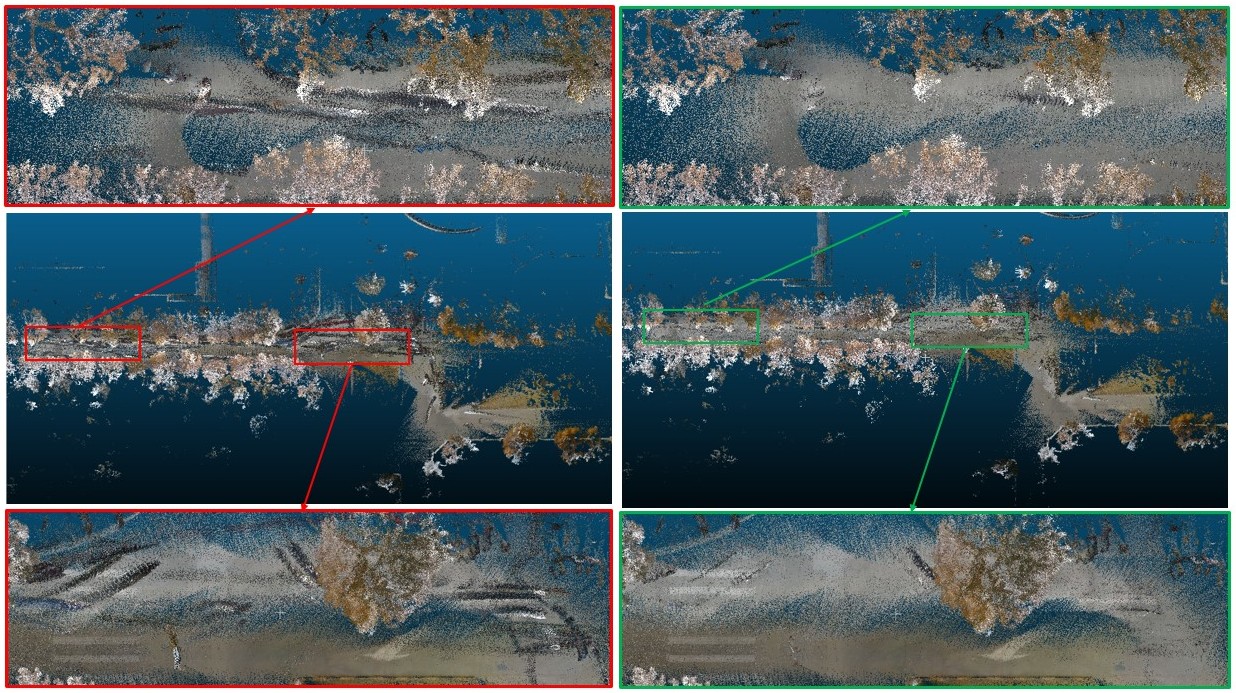}}
    \caption{Comparison of colored mapping results on M3DGR sequences.}
    \label{fig:M3DGRMapping}
\end{figure}
Fig.~\ref{fig:M3DGRMapping} shows the colored mapping results of FAST-LIVO2 and Dynamic-LIVO on two M3DGR dynamic sequences. The first sequence represents an indoor scenario in which several pedestrians continuously move around the room. As a result, the map generated by FAST-LIVO2 contains noticeable residual dynamic points corresponding to the moving pedestrians, whereas Dynamic-LIVO effectively removes most of these points and produces a cleaner static map. Nevertheless, a small number of residual dynamic points remain in the Dynamic-LIVO map. These residual points mainly arise when pedestrians briefly stop moving. During such short stationary periods, the temporal component derived from the S-T normal falls below the dynamic detection threshold, causing the corresponding points to be classified as static and retained in the map.

The second sequence represents a large-scale outdoor scenario with a long trajectory, where multiple pedestrians and vehicles move through the environment. The FAST-LIVO2 map exhibits noticeable ghosting artifacts caused by these moving objects. In contrast, Dynamic-LIVO effectively filters out most of the residual dynamic points, substantially reducing the ghosting artifacts and producing a cleaner static map. These results demonstrate the effectiveness of the proposed dynamic point filtering method and further show that Dynamic-LIVO is capable of constructing clean and dense colored static maps in real time across both indoor and outdoor dynamic environments.

\subsubsection{Self-Collected Datasets}
\begin{figure}
\vspace{2mm}
    \centering
    \subfloat[VR Room]{\includegraphics[width= 1.0\linewidth]{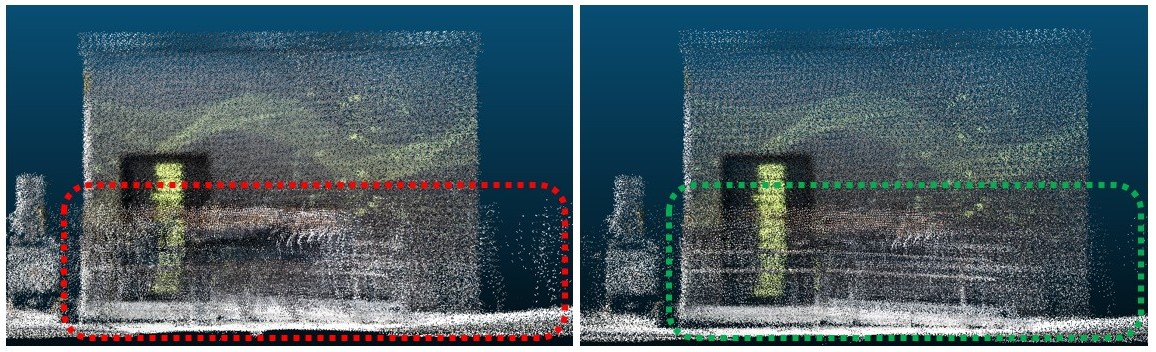}}
    \\ 
    \subfloat[Corridor]{\includegraphics[width= 1.0\linewidth]{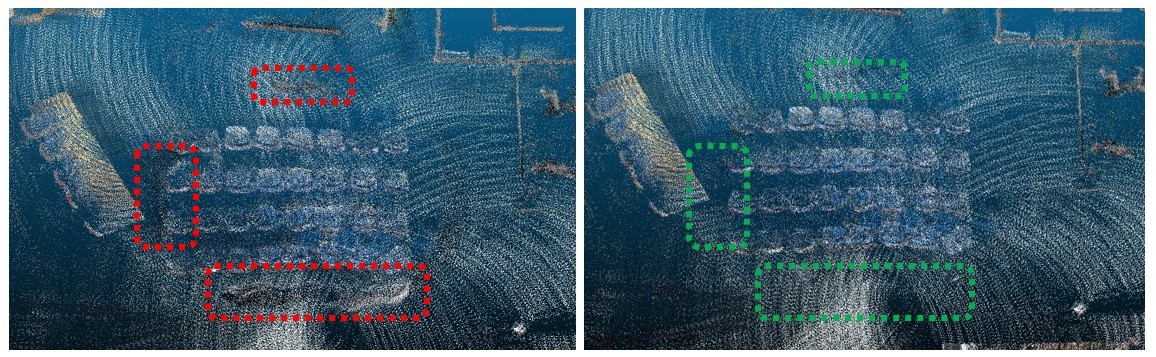}}
    \caption{Comparison of colored mapping results generated by FAST-LIVO2 (left) and Dynamic-LIVO (right) on the self-collected dynamic sequences. Red boxes highlight dynamic pedestrians retained in the maps generated by FAST-LIVO2, while green boxes indicate the corresponding regions after dynamic point removal by Dynamic-LIVO.}
    \label{fig:SelfMapping}
\end{figure}
Fig.~\ref{fig:SelfMapping} presents the colored mapping results of FAST-LIVO2 and Dynamic-LIVO on two representative sequences from our self-collected dataset. In the first scenario, two pedestrians repeatedly walk back and forth in front of a wall during data collection. Consequently, the FAST-LIVO2 map contains noticeable residual dynamic points, forming clear ghosting artifacts of the moving pedestrians and partially obscuring the wall behind them. In contrast, Dynamic-LIVO effectively filters out most of these dynamic points, substantially reducing the pedestrian artifacts and clearly revealing the static wall structure.

Similarly, in the second corridor scenario, two pedestrians move through the environment during data collection. The FAST-LIVO2 map exhibits noticeable residual dynamic points corresponding to the moving pedestrians, whereas Dynamic-LIVO effectively removes most of these points and reveals the static ground surface underneath.

\subsection{Dynamic Point Detection}
\begin{table}[t]
\vspace{2mm}
    \centering
    \caption{Quantitative evaluation of dynamic point filtering on
    self-collected sequences.}
    \label{tab:dynamic_filtering}
    \resizebox{\columnwidth}{!}{
    \begin{tabular}{lccccc}
        \toprule
        \multirow{2}{*}{Sequence} &
        \multicolumn{2}{c}{Dynamic-LIVO (w/o TD)} &&
        \multicolumn{2}{c}{Dynamic-LIVO} \\
        \cmidrule(lr){2-3}
        \cmidrule(lr){5-6}
        & SRR (\%) $\downarrow$
        & RNR (\%) $\downarrow$
        &&
        SRR (\%) $\downarrow$
        & RNR (\%) $\downarrow$ \\
        \midrule
        Office Fixed      & 2.35 & \textbf{8.35} && \textbf{1.10} & 8.77 \\
        Corridor Fixed    & 7.27 & 18.68 && \textbf{4.61} & \textbf{18.43} \\
        Poster Wall Fixed & 6.53 & 6.29 && \textbf{3.03} & \textbf{5.74} \\
        \bottomrule
    \end{tabular}
    }

    \vspace{1mm}
    \parbox{\columnwidth}{\footnotesize
    \raggedright
    * The best results are shown in \textbf{bold}.}
\end{table}

To quantitatively evaluate dynamic point filtering performance and validate the effectiveness of the proposed time-delayed S-T normal estimation, we conducted additional experiments in three self-collected scenes. For each scene, the sensor platform was rigidly fixed at the same position and orientation, and two LiDAR sequences of approximately 10s were recorded under static and dynamic conditions, respectively. The static sequence was processed by FAST-LIVO2 to construct a reference map, avoiding the removal of static points caused by false positives in S-T normal-based dynamic detection. The dynamic sequence, containing moving pedestrians, was then processed by Dynamic-LIVO and its variant without the time-delayed strategy (denoted as w/o TD). Keeping the sensor platform fixed eliminates the influence of localization errors, enabling a direct comparison of dynamic filtering performance.


We evaluate the filtering performance using two voxel-based metrics: the \emph{Static Removal Rate} (SRR) and the \emph{Residual Non-static Rate} (RNR). SRR measures the proportion of voxels in the static reference map that cannot be matched in the filtered result, reflecting static structures that are incorrectly removed during dynamic filtering. In contrast, RNR measures the proportion of voxels remaining in the result that cannot be matched to the static reference map, which mainly correspond to residual dynamic points. Therefore, lower SRR and RNR indicate better static point preservation and more effective dynamic point removal, respectively.

As shown in Tab.~\ref{tab:dynamic_filtering}, incorporating the time-delayed
S-T normal estimation strategy substantially reduces the SRR across all three
sequences, indicating that it effectively recovers static points that would
otherwise be falsely classified as dynamic due to insufficient spatio-temporal
observations. Meanwhile, the RNR exhibits only minor changes, demonstrating
that the proposed strategy improves static point preservation while maintaining
comparable dynamic point removal performance.

\subsection{Computational Time Analysis}
To evaluate the real-time performance of Dynamic-LIVO, we measured the execution time of each major module on the Corridor sequence of our self-collected dataset. The dynamic point classification and filtering pipeline requires an average processing time of 36.22~ms per frame. Among its major components, ST-normal computation and short-term dense map updating account for the majority of the computational cost, requiring an average of 5.42~ms and 23.45~ms per frame, respectively. In addition, the LIO and VIO subsystem updates require an average of 9.53~ms and 4.30~ms per frame, respectively. Consequently, the overall processing time of Dynamic-LIVO is approximately 50.05~ms per frame, which is well below the 100~ms interval between consecutive LiDAR scans at 10~Hz. These results demonstrate that Dynamic-LIVO can operate in real time while performing dynamic point filtering, state estimation, and static mapping.

\section{CONCLUSION}
In this paper, we proposed Dynamic-LIVO, a dynamic-aware LiDAR-inertial-visual odometry system for robust state estimation and static colored mapping in dynamic environments. By leveraging S-T normal analysis, Dynamic-LIVO identifies and filters dynamic LiDAR points and further excludes their associated visual observations from visual updates, reducing the influence of moving objects on both LIO and VIO. Experiments on public and self-collected datasets demonstrate improved localization accuracy and cleaner static colored mapping in challenging dynamic environments. Future work will focus on improving S-T normal-based dynamic point filtering and leveraging the constructed static colored maps for dense static scene reconstruction using 3D Gaussian Splatting.




\bibliographystyle{IEEEtran}
\bibliography{references}

@article{mur2017orb,
  title={Orb-slam2: An open-source slam system for monocular, stereo, and rgb-d cameras},
  author={Mur-Artal, Raul and Tard{\'o}s, Juan D},
  journal={IEEE transactions on robotics},
  volume={33},
  number={5},
  pages={1255--1262},
  year={2017},
  publisher={IEEE}
}

@article{qin2018vins,
  title={Vins-mono: A robust and versatile monocular visual-inertial state estimator},
  author={Qin, Tong and Li, Peiliang and Shen, Shaojie},
  journal={IEEE transactions on robotics},
  volume={34},
  number={4},
  pages={1004--1020},
  year={2018},
  publisher={IEEE}
}

@inproceedings{shan2020lio,
  title={Lio-sam: Tightly-coupled lidar inertial odometry via smoothing and mapping},
  author={Shan, Tixiao and Englot, Brendan and Meyers, Drew and Wang, Wei and Ratti, Carlo and Rus, Daniela},
  booktitle={2020 IEEE/RSJ international conference on intelligent robots and systems (IROS)},
  pages={5135--5142},
  year={2020},
  organization={IEEE}
}

@article{xu2022fastlio2,
  title={FAST-LIO2: Fast Direct LiDAR-Inertial Odometry},
  author={Xu, Wei and Cai, Yixi and He, Dongjiao and Lin, Jiarong and Zhang, Fu},
  journal={IEEE Transactions on Robotics},
  volume={38},
  number={4},
  pages={2053--2073},
  year={2022}
}

@article{zheng2024fast,
  title={Fast-livo2: Fast, direct lidar--inertial--visual odometry},
  author={Zheng, Chunran and Xu, Wei and Zou, Zuhao and Hua, Tong and Yuan, Chongjian and He, Dongjiao and Zhou, Bingyang and Liu, Zheng and Lin, Jiarong and Zhu, Fangcheng and others},
  journal={IEEE Transactions on Robotics},
  volume={41},
  pages={326--346},
  year={2024},
  publisher={IEEE}
}

@inproceedings{lichtenfeld2024efficient,
  title={Efficient dynamic LiDAR odometry for mobile robots with structured point clouds},
  author={Lichtenfeld, Jonathan and Daun, Kevin and von Stryk, Oskar},
  booktitle={2024 IEEE/RSJ International Conference on Intelligent Robots and Systems (IROS)},
  pages={10137--10144},
  year={2024},
  organization={IEEE}
}

@article{jia2025trlo,
  title={Trlo: An efficient lidar odometry with 3-d dynamic object tracking and removal},
  author={Jia, Yanpeng and Wang, Ting and Cao, Fengkui and Chen, Xieyuanli and Shao, Shiliang and Liu, Lianqing},
  journal={IEEE Transactions on Instrumentation and Measurement},
  volume={74},
  pages={1--10},
  year={2025},
  publisher={IEEE}
}

@article{shi2024dynam,
  title={Dynam-LVIO: A dynamic-object-aware LiDAR visual inertial odometry in dynamic urban environments},
  author={Shi, Jian and Wang, Wei and Qi, Mingyang and Li, Xin and Yan, Ye},
  journal={IEEE Transactions on Instrumentation and Measurement},
  volume={73},
  pages={1--19},
  year={2024},
  publisher={IEEE}
}

@article{chen2025breaking,
  title={Breaking the static assumption: a dynamic-aware lio framework via spatio-temporal normal analysis},
  author={Chen, Zhiqiang and Le Gentil, Cedric and Lin, Fuling and Lu, Minghao and Qiao, Qiyuan and Xu, Bowen and Qi, Yuhua and Lu, Peng},
  journal={IEEE Robotics and Automation Letters},
  year={2025},
  publisher={IEEE}
}

@inproceedings{le2024real,
  title={Real-time truly-coupled lidar-inertial motion correction and spatiotemporal dynamic object detection},
  author={Le Gentil, Cedric and Falque, Raphael and Vidal-Calleja, Teresa},
  booktitle={2024 IEEE/RSJ International Conference on Intelligent Robots and Systems (IROS)},
  pages={12565--12572},
  year={2024},
  organization={IEEE}
}

@article{falque2023dynamic,
  title={Dynamic object detection in range data using spatiotemporal normals},
  author={Falque, Raphael and Gentil, Cedric Le and Sukkar, Fouad},
  journal={arXiv preprint arXiv:2310.13273},
  year={2023}
}

@inproceedings{zhang2025towards,
  title={Towards robust sensor-fusion ground SLAM: A comprehensive benchmark and a resilient framework},
  author={Zhang, Deteng and Zhang, Junjie and Sun, Yan and Li, Tao and Yin, Hao and Xie, Hongzhao and Yin, Jie},
  booktitle={2025 IEEE/RSJ International Conference on Intelligent Robots and Systems (IROS)},
  pages={8894--8901},
  year={2025},
  organization={IEEE}
}

@inproceedings{lin2022r,
  title={R 3 LIVE: A Robust, Real-time, RGB-colored, LiDAR-Inertial-Visual tightly-coupled state Estimation and mapping package},
  author={Lin, Jiarong and Zhang, Fu},
  booktitle={2022 international conference on robotics and automation (ICRA)},
  pages={10672--10678},
  year={2022},
  organization={IEEE}
}

@inproceedings{shan2021lvi,
  title={Lvi-sam: Tightly-coupled lidar-visual-inertial odometry via smoothing and mapping},
  author={Shan, Tixiao and Englot, Brendan and Ratti, Carlo and Rus, Daniela},
  booktitle={2021 IEEE international conference on robotics and automation (ICRA)},
  pages={5692--5698},
  year={2021},
  organization={IEEE}
}

@inproceedings{zheng2022fast,
  title={FAST-LIVO: Fast and tightly-coupled sparse-direct LiDAR-inertial-visual odometry},
  author={Zheng, Chunran and Zhu, Qingyan and Xu, Wei and Liu, Xiyuan and Guo, Qizhi and Zhang, Fu},
  booktitle={2022 IEEE/RSJ international conference on intelligent robots and systems (IROS)},
  pages={4003--4009},
  year={2022},
  organization={IEEE}
}

@inproceedings{yuan2025lidar,
  title={LiDAR-inertial odometry in dynamic driving scenarios using label consistency detection},
  author={Yuan, Zikang and Wang, Xiaoxiang and Wu, Jingying and Cheng, Junda and Yang, Xin},
  booktitle={2025 IEEE/RSJ International Conference on Intelligent Robots and Systems (IROS)},
  pages={1598--1605},
  year={2025},
  organization={IEEE}
}

@misc{grupp2017evo,
  title={evo: Python package for the evaluation of odometry and SLAM.},
  author={Grupp, Michael},
  howpublished={\url{https://github.com/MichaelGrupp/evo}},
  year={2017}
}

@inproceedings{zuo2019lic,
  title={Lic-fusion: Lidar-inertial-camera odometry},
  author={Zuo, Xingxing and Geneva, Patrick and Lee, Woosik and Liu, Yong and Huang, Guoquan},
  booktitle={2019 IEEE/RSJ International Conference on Intelligent Robots and Systems (IROS)},
  pages={5848--5854},
  year={2019},
  organization={IEEE}
}

@inproceedings{zuo2020lic,
  title={Lic-fusion 2.0: Lidar-inertial-camera odometry with sliding-window plane-feature tracking},
  author={Zuo, Xingxing and Yang, Yulin and Geneva, Patrick and Lv, Jiajun and Liu, Yong and Huang, Guoquan and Pollefeys, Marc},
  booktitle={2020 IEEE/RSJ International Conference on Intelligent Robots and Systems (IROS)},
  pages={5112--5119},
  year={2020},
  organization={IEEE}
}

@article{bell1993iterated,
  title={The iterated Kalman filter update as a Gauss-Newton method},
  author={Bell, Bradley M and Cathey, Frederick W},
  journal={IEEE Transactions on Automatic Control},
  volume={38},
  number={2},
  pages={294--297},
  year={1993},
  publisher={IEEE}
}

@article{yuan2024sr,
  title={SR-LIVO: LiDAR-inertial-visual odometry and mapping with sweep reconstruction},
  author={Yuan, Zikang and Deng, Jie and Ming, Ruiye and Lang, Fengtian and Yang, Xin},
  journal={IEEE Robotics and Automation Letters},
  volume={9},
  number={6},
  pages={5110--5117},
  year={2024},
  publisher={IEEE}
}

@inproceedings{ester1996density,
  title={A Density-Based Algorithm for Discovering Clusters in Large Spatial Databases with Noise},
  author={Ester, Martin and Kriegel, Hans-Peter and Sander, J{\"o}rg and Xu, Xiaowei},
  booktitle={Proceedings of the Second International Conference on Knowledge Discovery and Data Mining},
  pages={226--231},
  year={1996}
}

@article{lin2024r,
  title={R 3 LIVE++: A Robust, Real-Time, Radiance Reconstruction Package With a Tightly-Coupled LiDAR-Inertial-Visual State Estimator},
  author={Lin, Jiarong and Zhang, Fu},
  journal={IEEE Transactions on Pattern Analysis and Machine Intelligence},
  volume={46},
  number={12},
  pages={11168--11185},
  year={2024},
  publisher={IEEE}
}

@inproceedings{chen2019suma++,
  title={Suma++: Efficient lidar-based semantic slam},
  author={Chen, Xieyuanli and Milioto, Andres and Palazzolo, Emanuele and Giguere, Philippe and Behley, Jens and Stachniss, Cyrill},
  booktitle={2019 IEEE/RSJ international conference on intelligent robots and systems (IROS)},
  pages={4530--4537},
  year={2019},
  organization={IEEE}
}

@inproceedings{henein2020dynamic,
  title={Dynamic SLAM: The need for speed},
  author={Henein, Mina and Zhang, Jun and Mahony, Robert and Ila, Viorela},
  booktitle={2020 IEEE International Conference on Robotics and Automation (ICRA)},
  pages={2123--2129},
  year={2020},
  organization={IEEE}
}

@inproceedings{Schmid2024Khronos,
title = "Khronos: A Unified Approach for Spatio-Temporal Metric-Semantic SLAM
in Dynamic Environments",
author = {Lukas Schmid and Marcus Abate and Yun Chang and Luca Carlone},
year = {2024},
booktitle = "Proc. of Robotics: Science and Systems"
}

@inproceedings{postica2016robust,
  title={Robust moving objects detection in lidar data exploiting visual cues},
  author={Postica, Gheorghii and Romanoni, Andrea and Matteucci, Matteo},
  booktitle={2016 IEEE/RSJ International Conference on Intelligent Robots and Systems (IROS)},
  pages={1093--1098},
  year={2016},
  organization={IEEE}
}

\end{document}